# ONTOLOGY ENABLED HYBRID MODELING AND SIMULATION

John Beverley[1] and Andreas Tolk[2]

[1]Dept. of Philosophy, University at Buffalo, Buffalo, NY, USA
Orcid: 0000-0002-1118-1738
[2]The MITRE Corporation, Charlottesville, VA, USA
Orcid: 0000-0002-4201-8757

**ABSTRACT**

We explore the role of ontologies in enhancing hybrid modeling and simulation through improved semantic rigor, model reusability, and interoperability across systems, disciplines, and tools. By distinguishing between methodological and referential ontologies, we demonstrate how these complementary approaches address interoperability challenges along three axes: Human–Human, Human–Machine, and Machine–Machine. Techniques such as competency questions, ontology design patterns, and layered strategies are highlighted for promoting shared understanding and formal precision. Integrating ontologies with Semantic Web Technologies, we showcase their dual role as descriptive domain representations and prescriptive guides for simulation construction. Four application cases – sea-level rise analysis, Industry 4.0 modeling, artificial societies for policy support, and cyber threat evaluation – illustrate the practical benefits of ontology-driven hybrid simulation workflows. We conclude by discussing challenges and opportunities in ontology-based hybrid M&S, including tool integration, semantic alignment, and support for explainable AI.

## 1 INTRODUCTION

This paper provides an overview of various ontology methods that enable the efficient use of hybrid modeling and simulation (M&S) by describing the hybrid components in a rigorous and formal way. Leveraging ontologies – controlled vocabularies representing entities in a given domain and logical relationships among them (Arp et al. 2015) – in support of M&S is not a new topic. Fishwick and Miller (2004) are among the first to show possible applications of ontology methods. Their main focus was to describe the simulation part, and how to capture the many aspects of developing a simulation and applying it. A more recent summary was provided by McGinnis et al. (2011).

These papers largely provide conceptualizations of the simulation paradigm, addressing the question of *HOW* something is modeled, which is the methodological perspective. While these aspects to align simulations as implementations of methods are important and necessary, they are not sufficient. As identified in multiple articles, such as (Tolk et al. 2012; Tolk 2018), it is crucial to also address the question of *WHAT* is modeled, which is the referential perspective of the conceptualization. This is of particular interest for the hybrid M&S community. One of the first papers looking into using ontologies to support conceptual alignment of models and interoperability of simulation evaluations for hybrid solutions was published by Yilmaz (2007).

As discussed in a panel on the purpose and benefits of hybrid simulation (Mustafee et al. 2017), *a hybrid is the result of merging two or more components of different categories to generate something new; something that combines the characteristics of these components into something more useful.* Hybrids enable the cognitive diversity needed to allow one to represent various facets, in the interest of understanding complex problems and contribute to their solutions. Enabling such cognitive diversity is the hallmark of interdisciplinary collaboration, as it contributes to better solutions by addressing the various aspects that are often only known to a subset of all participating experts. Hybrid modeling allows one to support the common conceptualization understandable by all participating experts, and hybrid simulation allows one to compose their various computational tools for collaboration. These ideas are described in more detail in (Tolk et al. 2021).

These two views of concept and implementation are mutually dependent. Simulations are dynamic representations of systems of interest within their environments, capturing how systems evolve over time, respond to events, and interact with surroundings. Various modeling paradigms – such as discrete event modelings, agent-based modeling, and system dynamics – as well as the modeling types – such as using ordinary differential equations, process algebra, and temporal logic – offer different approaches to representing such changes. However, these changes occur within the representation of the system and its environment, which is addressed through referential conceptualization.

Similar observations have been made within ontology engineering communities, where highly general ontologies serve dual roles, providing both referential and methodological conceptualizations within a single scope. One of the foundational articles addressing methodological and referential challenges of M&S and the role ontologies play is Hofmann et al. (Hofmann et al. 2011) which introduces the idea of separating between methodological and referential ontologies to address both aspects of conceptualization of the problem and implementation of these concepts as a computer simulation.

Recent years have witnessed significant improvements in the standardization of ontology development, computational support, and deployment strategies. Organizations are rapidly outgrowing the need for solutions limited to shared vocabularies, conceptual models, taxonomies, important as such products are. There is now a growing hunger for ontologies – and knowledge graphs constructed from them – that make explicit the deep implicit semantics buried within data. By providing a common formal framework, ontologies offer foundations for integrating knowledge from disparate domains, facilitate data exchange, and data reuse across context and operational units. Furthermore, the adoption of layered approaches – where upper ontologies represent highly general entities, domain ontologies focus on subject-specific concepts, and mid-level ontologies bridge the gap between these two levels – has promoted greater collaboration, conceptual reuse, and even standardization (Beverley et al. 2024). Illustrative examples are ontologies developed under the top-level Basic Formal Ontology (BFO) (Otte et al. 2022), ISO/IEEE Standard No. 21838-2:2020 (ISO/IEC 2021b), which is used in more than 700 open-source ontology projects ranging from biology (Smith et al. 2007) and industrial manufacturing (Drobnjakovic et al. 2022), to defense and intelligence (Wade and Martell 2024).

Here we provide examples of modern ontological techniques used to support the unambiguous representation of methodological and referential aspects of hybrid M&S. In doing so, we advocate for the use of such methods to improve hybrid M&S research and results in the future.

## 2 ONTOLOGICAL FOUNDATIONS

Central to ontology engineering is the promotion of interoperability: the ability of systems, organizations, or agents to exchange and meaningfully use information. Despite the importance of interoperability, the term itself is highly ambiguous (Wegner 1996; Curts and Campbell 1999; Tolk 2013; Ford et al. 2007; Berre et al. 2007). Without engaging in definitional debates, ontologists distinguish three distinct axes of interoperability, each with its own goals and evaluative criteria: Human-Human, Human-Machine, and Machine-Machine Interoperability. Challenges encountered along each axis have analogues to challenges in hybrid M&S; ontology engineering strategies for addressing such challenges may also have analogues in M&S, as discussed in detail in (Tolk et al. 2021).

The first axis – Human-Human Interoperability – concerns the alignment of understanding between human agents. Ontology development facilitates Human–Human Interoperability by formalizing domain knowledge through consensus-driven discussions between ontologists and domain experts, to identify shared semantic structures (Neuhaus and Hastings 2022). Two practices are commonly deployed along this axis: the use of competency questions (Keet 2020) and ontology design patterns (Carriero 2024).

Competency questions are natural language prompts that reflect questions an ontology should be designed to answer, serving both as requirements and for validation. For example, in a logistics-focused simulation, a competency question might be: "Which resources are assigned to emergency zones after a flood event?" or "What delivery routes remain valid after a distribution center becomes inoperable?" Such questions clarify the scope of the ontology, surface implicit assumptions, and help align contributors from different disciplines by explicitly stating what knowledge the ontology must encode. In multi-stakeholder modeling scenarios – common in hybrid M&S – competency questions guide elicitation of domain knowledge while minimizing ambiguity, creating a shared design space, and supporting alignment goals. Analogous to software design patterns, ontology design patterns reflect best practices for representing recurring semantic structures, such as participation in events, part-whole relationships, roles, and temporal constraints. For example, the "n-ary relationship" pattern allows a designer to model situations where more than two entities are related – such as a person playing a role in an organization during a time interval (Natasha Noy and Alan Rector 2006). In hybrid M&S, patterns are especially valuable in structuring interactions across model types coupling social, technical, and environmental perspectives – such as agent behaviors influencing system dynamics flows and domains.

Human–Machine Interoperability concerns the ability of computational systems to interpret and act upon human-understandable descriptions of the world. Bridging this gap requires translating domain knowledge into formal representations that machines can reason over, such as ontologies, logical models, schema, and so on. Along this axis, ontology engineers often adopt a layered modeling strategy that reflects degrees of semantic generality. At the most general level, "top-level" ontologies define abstract categories – such as object, process, role, location, and time – that serve as shared anchors for more specific domain knowledge (ISO/IEC 2021a). Mid-level ontologies then build on this foundation to capture cross-cutting patterns relevant across multiple domains – like "infrastructure component" or "health event" – reducing redundancy and promoting reuse (Pease and Benzmueller 2010; Beverley et al. 2024). Finally, domain ontologies express specialized vocabularies, relationships, and distinctions that arise in specific fields of interest, such as emergency logistics, climate forecasting, public health, and so on (Arp et al. 2015).

A concrete implementation of the layered approach can be seen in BFO (Otte et al. 2022), the top-level ontology from which the Common Core Ontologies (CCO) mid-level suite of ontologies extend, which in turn provide a foundation for various domain-level ontologies (Jensen et al. 2024), much like Tensorflow and Pytorch extend from a more general Python language. Every root term in each of the 11 OWL files that comprise the CCO suite, extends from a leaf term found in BFO following a recipe to ensure there remains a common semantics as terms are populated downward. For example, the CCO term "agent" is defined as a (BFO) material entity capable of engaging in some intentional action. The recipe on display in this example is that a given class 'A' is defined according to the pattern "is a B that Cs" where 'B' is the parent class of 'A', and 'Cs' designates features of 'A' that distinguish it from all other 'Bs'.

Definitions are encoded in the World Wide Web Consortium (W3C) Web Ontology Language 2 (OWL2) (W3C 2009), an extension of the Resource Description Framework (RDF) and RDF Schema (RDFs), which require data be represented as sets of subject-predicate-object directed graphs (W3C 2004). RDF and RDFs languages provide a vocabulary for representing classes of entities, such as tables and chairs, as well as instances of such classes, such as this table and these chairs. OWL2 extends these vocabularies to allow for representation of logical relationships among classes and among instances, such as when two classes are disjoint from one another, i.e. have no instances in common. Formally, the OWL2 vocabulary is a decidable language, supporting automated consistency and satisfiability checking (Baader et al. 2017) using automated reasoners such as HermiT (Shearer et al. 2008) and Pellet (Sirin et al. 2007).

The third axis – Machine–Machine Interoperability – focuses on enabling autonomous systems to exchange, interpret, and reason over information without human intervention. Whereas Human–Human and Human–Machine Interoperability deal with aligning semantics between people and between people and machines respectively, Machine–Machine Interoperability hinges on machines having access to semantically coherent, logically rigorous, and operationally actionable knowledge representations. Ontologies along this axis act as interface contracts between independent components, allowing for seamless integration across disparate simulation engines, data pipelines, or autonomous agents within a hybrid M&S framework.

## 3 ONTOLOGY FOR HYBRID M&S

### 3.1 Methodological Ontologies

Methodological ontology methods encompass the systematic development and implementation of formal ontologies that facilitate interoperability across how simulations are modeled. DeMO, which is an ontology for discrete-event M&S (Silver et al. 2011), is a well-known example. Their components are based on the use of state-oriented, event-oriented, activity-oriented, and process-oriented models. First steps towards capturing hybrid simulation ontology components are described in (Saleh, Bell, and Sulaiman 2021). For the objective of this paper, however, it is not so important which of the modeling paradigms and methods is used, but that a methodological ontology is a blueprint on how to model something using the chosen paradigm. This is a normative use of an ontology, which needs to be precise and unequivocal (Hofmann et al. 2011).

This normative role becomes especially significant when viewed through the lens of the three axes of interoperability. On the Human–Human Interoperability axis, methodological ontologies serve to align understanding between modelers from different disciplines by establishing a formal vocabulary for describing modeling assumptions and simulation constructs. This can be facilitated by leveraging competency questions to help scope and validate methodological ontologies involved in hybrid M&S efforts, where integration often requires resolving conceptual approaches to time, state, behavior, and causality.

On the Human–Machine Interoperability axis, methodological ontologies transition from communication frameworks to layered executable design grammars. The layered approach to ontology engineering enables this transition by distinguishing encodings of top-level categories, mid-level modeling strategies, and domain-level simulation modeling constructs. Participants can thus contribute at the level most suited to their expertise while maintaining semantic coherence across the broader modeling effort. Throughout, each layer can be represented in OWL2 and extended with rule-based formalisms like the Semantic Web Rule Language (SWRL) (Horrocks, Patel-Schneider, Boley, Tabet, Grosof, Dean, et al. 2004) and the Shapes Constraint Language (SHACL) (Gayo et al. 2018), enabling machines to parse, validate, and enforce simulation specifications. For example, they may specify which components are mandatory in a valid model, which parameter constraints must be satisfied, or how behaviors should be triggered under certain conditions. In hybrid simulation environments, such logic-driven structure supports the automatic generation of simulation code, the validation of simulation integrity, and the on-the-fly composition or adaptation of models based on external conditions or inputs.

Finally, methodological ontologies support Machine–Machine Interoperability by acting as semantic contracts between independently developed simulation modules. In hybrid workflows combining agent based, discrete-event, and system dynamics models, methodological ontologies prescribe how concepts like entities, events, and dependencies are to be instantiated and interpreted across tools. This ensures that independently built models can be integrated without violating their internal logic or misaligning their execution semantics. Ontologies that function as meta-models – such as the Simulation Ontology developed by (Jurasky et al. 2021) – demonstrate this function by enabling automated translation from conceptual design to executable simulation artifacts across modeling paradigms.

In each case, the methodological ontology serves as more than a descriptive vocabulary: it is a normative artifact that defines how simulation models are to be constructed, interpreted, and evaluated across users, interfaces, and machines.

*3.2 Referential Ontologies*

Referential ontology methods encompass the systematic development and implementation of formal ontologies that facilitate interoperability across what is modeled. In contrast to the methodological ontologies, referential ontologies are descriptive and capture pieces of the semantic relations of the real world, as they are understood by the modeler (Hofmann et al. 2011). As discussed in (Tolk et al. 2013), this can lead to inconsistencies in the description when all possible viewpoints are captured, as viewpoints are not necessarily conceptually aligned. The referential ontology now serves two purposes:

- It captures all viewpoints in a precise description of the system of interest, including all possible different views that may be mutually exclusive.
- It can be used as a reference model of the underlying knowledge that can then be tailored to consistent subsets of knowledge that can be implemented as a simulation.

Implementing all necessary simulations to use all available knowledge allows an extensive span of strategies for addressing any questions of interest regarding the modeled system.

At a general level, referential ontologies can be understood as precise descriptions of views of a system. This characterization captures both their formal nature—being structured, logic-based representations—and their perspectival quality—being anchored in specific conceptualizations that reflect a stakeholder's or discipline's orientation toward the system. As such, referential ontologies encode these perspectives in a computationally interpretable way, making them reusable across multiple contexts while supporting semantic clarity and alignment. This dual capacity – to both accommodate plural viewpoints and to support formalized semantic alignment – aligns closely with the three axes.

On the Human–Human axis, referential ontologies support communication between experts from distinct domains by providing common vocabularies and shared conceptual scaffolding. This is especially critical in hybrid M&S projects that involve stakeholders from engineering, policy, logistics, and other fields. Competency questions guide referential ontology development toward practical use cases and ensure mutual understanding of modeling goals. Complementing this, ontology design patterns provide representations of recurring patterns exhibited in the domain of interest, such as correlative relationships between engine parts and fuel, or cyber exploits and hardening techniques. Using these patterns across domains promotes high level semantic consistency while allowing explicit representation of semantic differences across specifications. Together, competency questions and design patterns help ensure that referential ontologies are both semantically expressive and collaboratively intelligible.

On the Human–Machine axis, referential ontologies in OWL2 facilitate simulation tools and reasoners to validate logical consistency, infer class relationships, and check constraint satisfaction. The layering strategy applies just as well to referential ontologies. Top- and mid-level ontologies provide foundations for domain level extensions, where highly specific domain ontologies extending from the same top- and mid-level may overlap in scope with other such domain ontologies. Aggregating ontologies with overlapping domains within layered structures invariably raise concerns of conflicts, but ontology engineers have strategies for managing such issues. On the one hand, ontologies can formally distinguish domain ontologies representing, say, "targets" as necessarily having material parts from domain ontologies that represent targets more flexibly, by encoding these semantic differences explicitly, even if the same label "target" is used for both. On the other hand, when domain ontologies overlap and exhibit mutually inconsistent structures, they are nevertheless connected by a common top- and mid-level ontology reflecting shared semantics. In this manner, layering supports the precise description of all viewpoints in a system, including those that may be mutually inconsistent.

On the Machine–Machine axis, referential ontologies enable distributed simulation components to exchange semantically coherent information. When simulation models developed in different environments or paradigms refer to the same underlying ontology, they can operate over shared formal structure, easing data transformations, as well as federated querying. For instance, if multiple agents in a federated simulation refer to the same concept of "supply chain bottleneck", an OWL2 encoded representation in a referential ontology leveraged by different systems will ensure that this term has a consistent formal definition, allowing different components to reason over it without ambiguity.

## 3.3 Ontologies for a General Modeling Framework

Tolk et al. (2013) introduced the M&S System Development Framework (MS–SDF) to guide development, as shown in Figure 1, exhibiting three main components: reference model, conceptual model, simulation model alongside its seven development steps, each of which can be integrated with ontological strategies aligned to the three axes of interoperability.

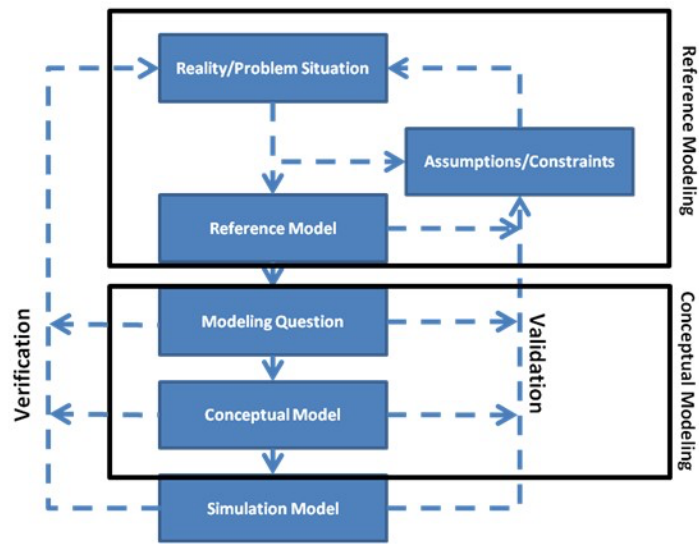

Figure 1: The M&S System Development Framework.

1. *Reality/Problem Situation:* Identify the phenomenon of interest where stakeholders have divergent views. Capture all key perspectives as true/false statements about the situation within a referential ontology (Human–Human). This establishes the foundation for the M&S process.
2. *Establish Assumptions/Constraints:* Represent stakeholder assumptions as explicit statements that can be evaluated as true/false, in the ontology (Human-Machine). Iterate through these assumptions, determining which to keep and which to discard, using a process to navigate the insight space for interdisciplinary teams (Shults and Wildman 2020). The result is a reference model combining knowledge with reasonable assumptions.
3. *Capture the Reference Model:* The reference model represents current knowledge and assumptions about the problem situation, which may be contradictory statements, as completeness in framing the problem is the priority (Human-Human, Human-Machine). This becomes the constructed reality upon which subsequent artifacts depend.
4. *Formulate a Modeling Question:* Develop specific queries about entities and their relationships as captured in the reference model. These questions determine whether the reference model contains the required elements to provide answers. Such questions can be designed to span the three axes of interoperability.
5. *Create the Conceptual Model:* Derive a model that answers the modeling question without violating assertions in the reference model. Resolve inconsistencies through subject matter expertise or additional constraints. The conceptual model must remain logically consistent to prevent unexpected behaviors in simulations, while balancing human perspectives (Human–Human) with formal structures (Human–Machine). If alternative mutual exclusive possibilities are part of the knowledge then multiple conceptual models will have to be derived to address all aspects captured in the reference model to build an ensemble.
6. *Create the Simulation Model:* Implement the conceptual model using appropriate software tools. The use of methodological ontologies should guide this process to ensure resulting simulations are precisely defined as the realization of the conceptual model, separate from the software components used to generate it. Finding balance between assumptions/constraints is crucial to maintain validity and ensure the model can be simulated (Machine-Machine).
7. *Verification and Validation:* Verification and validation become unified. Validation relates the simulation to the conceptual model, modeling question, reference model, and the assumptions as axioms, captured in the referential ontology. Verification examines relationships between these elements using the methodological ontology, where model checking tools facilitate validation against the reference model. This step ensures that the simulation accurately reflects stakeholder perspectives (Human–Human), adheres to formal specifications (Human–Machine), and can be integrated with other simulations (Machine–Machine).

With the increased use of ontologies to represent domain knowledge, the applicability of the M&S–SDF is becoming significantly easier than at the time it was developed. It combines the referential ontologies to capture what is or shall be modeled with the methodological ontologies describing the hybrid modeling and simulation methods that are or shall be applied. Moreover, by interpreting steps of the MS–SDF through the three axes of interoperability, we highlight the manner in which ontologies can support all critical facets of hybrid M&S: aligning human understanding, enabling machine interpretation, and facilitating seamless system integration.

# 4  APPLICATION EXAMPLES

Examples of creating hybrid simulations with ontologies are not as rare as one might think. In this section, we provide well documented applications to show feasibility and applicability of the methods proposed in the first half of the paper. For additional details we refer the interested reader to the original publications.

## 4.1 Use Case: Sea-level Rise Analysis

The first example utilized ontologies to represent 18 categories in the Department of Homeland Security Infrastructure Data Taxonomy, to provide stakeholders of the Hampton Roads, VA region a decision support simulation aimed to explore what areas should be evacuated or protected due to flooding 50 years from now in the event of a possible sea level rise (Tolk et al. 2013). It uses referential ontologies to align the views of these groups to allow to create a hybrid simulation. As the article states:

> The Hampton Roads region of southeastern Virginia is one of the most threatened areas in America. Most land within the region lies less than 15 feet above mean sea level and the combined impacts of rising waters and continental plate sink result in forecast relative sea level increases of 1½ to 2 feet over the next 100 years. Numerous studies have projected the extent of potentially flooded land areas and mapped threatened communities. Others have listed decision–impacting influences from particular perspectives. However, no studies have objectively identified a comprehensive list of factors that must be considered in the decision-making process and organized these factors in a manner that considers all simultaneously.

Interdisciplinary knowledge is provided using various conceptualizations that address the diverse views of the challenge from the perspectives of specialized experts. By aligning the conceptualizations using one common domain ontology to map all conceptualizations to, a computational decision support tool can be developed that represents all identified viewpoints. By capturing the assertions of the decision makers – such as their options for action, the costs of such actions, their goals, their terms of employment, and their resources constraints – different courses of actions were implemented from those assertions and the effects and costs visualized based on the common understanding of the challenge. This in turn aided the identification of what areas should be protected in accordance with all collected viewpoints, what protection will costs, as well as mitigating factors to consider.

Thehybridsimulationwasthendevelopedusingthealignedreferentialontologiestoderivetheconceptual model and its implementation following the steps of the MS–SDF described before.

## 4.2 Use Case: Industry 4.0 Modeling

The second example is more recent and proposes a framework on how to efficiently derive a simulation model from semantic knowledge bases (Jurasky et al. 2021). The authors propose a simulation ontology enabling a meta-model for hybrid simulations. Following the ideas captured in (Kumar et al. 2019), they use referential ontologies to describe the various views of the participating industry domains in *'the common language of an ontology.'* However, they also observe the need to support dynamic scenarios, for which they recommend the use of simulations, since widely used across manufacturing domains. They observe most simulation solutions are in fact software solution independent of knowledge representation standards, The separation of domain knowledge and simulation modeling expertise often results in long development times, which are contrary to the need for rapid agile decision making. Among others, they point to the theoretical concept for linking simulations to a semantic knowledge base by Rabe and Gocev (2012).

Their framework presents an ontology-based approach for transforming complex systems into executable simulations, anchored by a simulation ontology, which is a formal meta-model designed for hybrid simulations. This ontology provides a detailed class taxonomy and semantic relationships to capture key simulation elements, making them broadly applicable and understandable across various problem domains. Users can instantiate classes and properties within this ontology to formally define simulation models, marking a notable improvement in the conceptualization phase of simulation modeling.

To connect problem descriptions to simulation designs, the framework includes Mapping Rules that guide the translation of objects and constraints from a Use Case Ontology into the Simulation Ontology. A subsequent Parser then automatically generates executable code from this ontological representation, addressing the limitation that conceptual models lack direct experimental capability. For knowledge representation, the framework advocatesOWL2forrepresentationand SPARQL for querying. For execution, AnyLogic is recommended (Borshchev 2014), leveraging its XML-based AnyLogic Project (ALP) language and Java, particularly for its strength in hybrid modeling. The framework remains adaptable, allowing parser modifications to support other simulation software. This approach enhances simulation modeling with a semantically rich method for conceptualizing and implementing models across diverse domains. The original article (Jurasky et al. 2021) describes the simulation ontology in detail and even provides access to the online version provided as a data annex by the publisher.

*4.3 Use Case: Artificial Societies for Policy Support*

The third example uses artificial societies to provide a variety of options for better policy support (Clemen et al. 2025). Artificial societies extend the agent metaphor by introducing social nets as the third component to the representation of acting individuals as intelligent agents within their situated environment, which represents physical constraints for actions.

Ontologies play a pivotal role in this context, as they allow the machine-readable expression of the various belief systems and value systems of diverse social groups, which is required to conduct social research with such simulation representations of complex social-technical systems. They also allow the translation of social theories into actionable rule sets and algorithms for those agents who belong to the respective group. The importance of this has been exemplified in (Bullock et al. 2023), in which the authors used social science research results to apply fusion theory and moral foundations theory to generate behavior of agents of their artificial society in accordance with insights of experts in social sciences.

Like in the already identified challenge of supporting interdisciplinary research using hybrid modeling (Tolk et al. 2021), the different world views of the members of the various social nets are likely to require diverse and mutually supportive modeling approaches to make them computationally accessible, emphasizing the need to allow the representation of multiple different facets when dealing with challenges in complex social-technical systems.

Another aspect discussed by Clemen et al. (Clemen et al. 2025) is the need to exchange information between real world observations, artificial societies, and vice versa, enabling a social digital twin that observes the real world state, instantiates the elements of the artificial society, then provides services to the decision makers with actionable recommendations. Again, the heterogeneity of real world data as well as the multi-modality of methods and tools to provide recommendations can be complemented by the use of ontologies which capture aspects of both referential and methodological ontologies along all three axes of interoperability.

A final idea introduced by Clemen et al. (Clemen et al. 2025) that can be facilitated by ontologies is the use of generative Artificial Intelligence (AI) with focus on Large Language Models (LLM) to provide a language for the agents, but also to improve their behavior and other aspects of the artificial society, particularly when used as a societal digital twin. While many current publications focus on the use of LLMs to construct ontologies, the idea to use ontologies to inform LLMs better is a topic of growing interest. Referential and methodological ontologies provide structured knowledge about the object and its behavior that can support LLMs to reason and find solutions within these domain. An overview of current activities is presented in (Neuhaus 2023).

*4.4 Use Case: Cyber Threat Evaluation*

A fourth example applies ontology-based modeling to the defense of power monitoring systems against coordinated cyber-physical attacks (Teng et al. 2025). The authors introduce the Ontology of CyberPhysical Security for Power Monitoring System (OntoCPS4PMS), to support knowledge-driven simulation and reasoning for security analysis in critical infrastructure. Drawing from models such as MITRE's ATT&CK (Vanamala et al. 2020) and CAPEC (Strom et al. 2020), and aligning with the goals of D3FEND (Kaloroumakis and Smith 2021) for defense-centric modeling, they construct a multi-layered ontology that represents attackers, methods, vulnerabilities, and impacts in a unified semantic model.

At its core, OntoCPS4PMS combines static knowledge representation and dynamic inference by modeling entities and their interrelations across five dimensions: *Attacker*, *Attack Method*, *Target*, *Impact*, and *Security Defense*. Each dimension includes detailed subcategories, such as attacker motivation and resource level, software and malware tools, physical and cyber targets, strategic intent and consequences, as well as mitigation and defense solutions.

Simulation capabilities are implemented through SWRL (Horrocks, Patel-Schneider, Boley, Tabet, Grosof, Dean, et al. 2004), allowing for inference rules that simulate cyber-physical attack scenarios, assess system risk levels, and recommend appropriate mitigation measures. These rules function over OWL2 encodings, providing executable logic that mimics attacker behavior and system response. For example, one rule computes the risk severity of a security event by integrating vulnerability scores, attack likelihood, and tactic severity, captured in SWRL:

$$\text{Vulnerability}(?v) \wedge \text{CVSSscore}(?v,?s) \wedge \text{High}(?s) \wedge \text{Attack\_Pattern}(?ap) \wedge$$
$$\text{LikelihoodOfAttack}(?ap,?l) \wedge \text{High}(?l) \wedge \text{Tactic}(?t) \wedge \text{SeverityLevel}(?t,?sl) \wedge \text{High}(?sl)$$
$$\rightarrow \text{HighEvent}(?e)$$

This rule identifies an event as high-risk if it involves a highly severe tactic, an attack pattern with a high likelihood, and a vulnerability with a high CVSS score. Such logical constructs demonstrate how simulation logic—typically embedded in external tools—can instead be expressed and executed within an ontology itself, enabling semantically driven threat modeling. To validate the approach, the framework is applied to the 2015 Ukrainian power grid attack. Using real-world data from MITRE ATT&CK, CAPEC, and NVD, OntoCPS4PMS was able to simulate the intrusion process—from malware deployment to command issuance and data destruction—while recommending specific mitigation strategies.

By combining domain knowledge and simulation logic, OntoCPS4PMS illustrates how ontologies can unify both. It acts not only as a static ontology but also as a simulation driver, producing actionable insights through semantic inference. This hybrid approach enhances explainability, adaptability, and decision-support in rapidly evolving cyber-physical environments.

## 5  DISCUSSION

Adopting ontology engineering strategies supports M&S by addressing all aspects and facets at all abstraction levels throughout the life cycle. As hybrid M&S applications increasingly involve coupling knowledge across traditionally siloed domains—such as climate science and logistics, or cybersecurity and human behavior, layered ontological frameworks can be leveraged to support semantic bridging across disparate knowledge systems, while preserving internal coherence and inferential rigor. To reach this goal and provide the envisioned benefits, the conceptualizations of modeling types and paradigms used must be captured methodologically, and the diverse conceptualizations of the modeled system of interest by the participating groups must be captured referentially. In this paper, we propose the use of methodological and referential ontologies to accomplish this objective.

That said, there are challenges to overcome. First, ontology development, especially at scale, is resource intensive. While the layered approach provides modularity and reusability, the building, validating, and maintaining ontologies that are both expressive and computationally tractable often requires significant effort and expertise. Second, though it is clear that referential and methodological ontologies can be logically aligned, achieving such alignment in practice is non-trivial. Semantic mismatches between domain models and simulation paradigms can emerge, especially in interdisciplinary settings. The management of cross-layer dependencies, inconsistency detection, and conflict resolution across mappings is an area of on-going research in ontology engineering (Prudhomme et al. 2025). Third, the current ecosystem of tools supporting ontology-based hybrid M&S lacks tight integration with widely used modeling platforms. While standards like SWRL and SHACL offer expressive rule capabilities, they are not yet natively embedded in most simulation engines. Bridging this gap may require new interfaces or model compilers that interpret ontological constraints directly as executable logic.

We do not shy away from these challenges. Going forward, the development of integrated modeling environments for ontology editors, reasoners, and simulation engines should be a high priority. Similarly, as hybrid M&S models scale in complexity, modular design patterns and versioning strategies will be needed to keep ontologies adaptable and maintainable. Theoretical work on design patterns and top-, middle, and domain- architectures may lay the foundation for such practical work (Beverley et al. 2024). Additionally, building on recent interest in explainable AI and ontology-grounded learning, future research should explore how ontologies can support dynamic adaptation of simulation behavior in response to new data, goals, or external interventions.

As the scale, scope, and complexity of simulation-based research continues to grow, the case for interoperability-focused, layered ontology strategies to support hybrid M&S efforts becomes stronger. By distinguishing and coordinating referential and methodological layers, and aligning them with both human and machine interoperability requirements, ontologies provide the structure and semantic discipline needed to build simulations that are not only powerful and flexible, but also interpretable, reusable, and reliable. By advancing the integration of methodological and referential ontologies, this paper provides a foundational framework for enhancing hybrid modeling and simulation, paving the way for increasingly interoperable, adaptable, and insightful solutions to complex interdisciplinary challenges.

DISCLAIMER